%% file: main.tex
\documentclass[journal]{IEEEtran}

\usepackage{amsmath}
\usepackage{amssymb}
\usepackage{amsfonts}

\usepackage{graphicx}
\usepackage{booktabs}
\usepackage{array}
\usepackage{multirow}   %
\usepackage{placeins}   %
\usepackage{pifont}     %

\usepackage{url}
\usepackage[hidelinks]{hyperref}
\hypersetup{
  pdftitle={Omni-Woodpecker: Hallucination Correction via Evidence-Prior Rebalancing for Omni LLMs},
  pdfauthor={Zijian Liu, Chaoyou Fu, Yan Yang, Caifeng Shan, Ran He},
  pdfsubject={IEEE Transactions on Multimedia submission},
  pdfkeywords={Audio-visual hallucination, cross-modal conflict, omni-modal large language models, representation editing, training-free inference},
  pdfcreator={pdfLaTeX}
}

\title{Omni-Woodpecker: Hallucination Correction via Evidence-Prior Rebalancing for Omni LLMs}

\author{Zijian~Liu,
        Chaoyou~Fu,
        Yan~Yang,
        Caifeng~Shan,
        and~Ran~He,~\IEEEmembership{Fellow,~IEEE}%
\thanks{Zijian Liu, Chaoyou Fu, Yan Yang, and Caifeng Shan are with the School of
Intelligence Science and Technology, Nanjing University, Suzhou 215163, China.}%
\thanks{Ran He is with the University of Chinese Academy of Sciences, Beijing
100049, China.}%
\thanks{Corresponding author: Chaoyou Fu (E-mail: bradyfu24@gmail.com).}}%

\begin{document}
\bstctlcite{IEEEexample:BSTcontrol}

\maketitle

\begin{abstract}
Recent Omni LLMs (OLLMs) have demonstrated remarkable capabilities in joint audio-visual reasoning, yet cross-modal hallucinations remain a major challenge. Existing mitigation methods mainly operate at the output level, overlooking the entanglement of grounding evidence and learned priors within fused representations.
To address this issue, we propose Omni-Woodpecker (OWP), a training-free framework that shifts hallucination mitigation from output-level manipulation to conflict-conditioned representation rebalancing.
Specifically, OWP first identifies cross-modal conflicts from complementary modality-specific views.
It then employs orthogonal decomposition to separate grounding evidence from competing priors within the hidden states.
Finally, OWP adaptively rebalances fused representations by suppressing misleading priors and restoring ignored evidence before decoding.
Extensive experiments demonstrate the effectiveness and strong cross-architecture generalization of OWP.
Using \mbox{Qwen2.5-Omni-7B} as the backbone, OWP improves performance by 12.9 and 8.6 points on CMM and AVHBench, respectively, while consistently achieving gains of 0.67--2.25 points across four general OLLM reasoning benchmarks.
Furthermore, OWP generalizes effectively to \mbox{VideoLLaMA2-AV}, improving performance by 4.6 and 3.9 points on the same benchmarks. Code is publicly available at \url{https://github.com/ZiiJianLiu/Omni-Woodpecker}
\end{abstract}

\input{sec/01Introduction}
\input{sec/02relatedwork}
\input{sec/03method}
\input{sec/04experiments}

\input{sec/06limitations}
\input{sec/05conclusion}

\bibliographystyle{IEEEtran}
\bibliography{references}

\end{document}

%% file: sec/01Introduction.tex
\section{Introduction}

\begin{figure}[t]
\centering
\includegraphics[width=1.0\columnwidth]{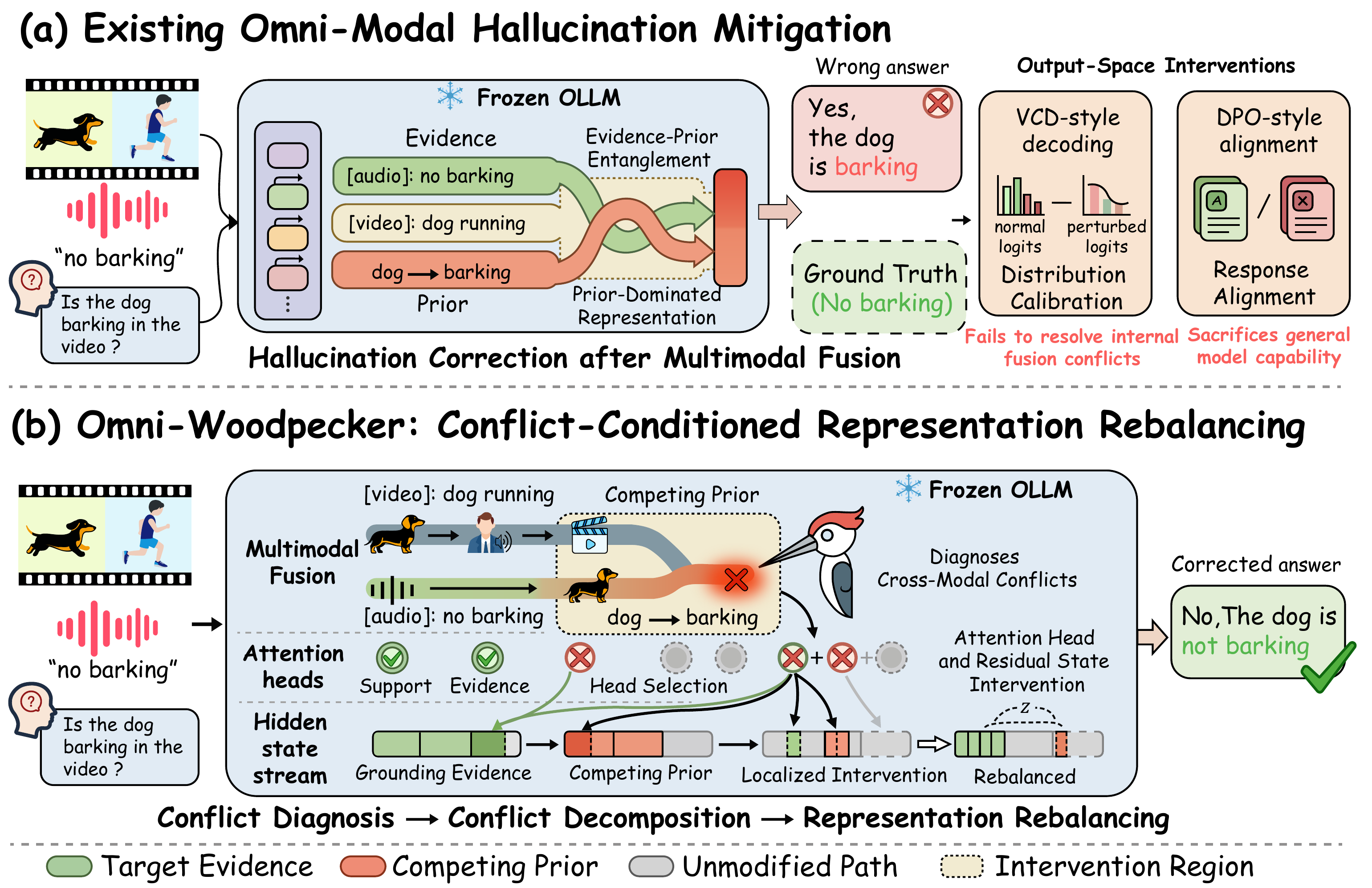} %
\caption{\textbf{Comparison of omni-modal hallucination mitigation paradigms.}
\textbf{(a)} Existing output-level methods correct hallucinations after multimodal fusion. They lack direct control over the evidence-prior interactions already encoded in the fused representations.
\textbf{(b)} Omni-Woodpecker (OWP) diagnoses conflicts using complementary evidence views. 
It orthogonally separates grounding evidence from competing priors, and adaptively restores their balance within the fused representations before decoding.
}
\label{fig1}
\end{figure}

Omni LLMs (OLLMs) extend LLMs beyond text by jointly processing language, vision, and audio~\cite{qwen25omni,qwen3omni,omnivinci}, enabling models to reason across complex environments. However, it introduces a reliability challenge: when different modalities provide incomplete or conflicting cues, the model struggles to appropriately balance grounding evidence against learned priors, and omni-modal hallucinations typically emerge when this balancing process fails.

As shown in Fig.~\ref{fig1}a, when asked ``Is the dog barking?'', the model actually encodes the evidence of silence in the audio. However, a strong visual prior (seeing a dog) overpowers this audio evidence during multimodal fusion. This shows that the model does not fail to perceive the inputs. Instead, competing cross-modal cues suppress the target evidence. This critical observation drives our core hypothesis: omni-modal hallucination is not a failure of evidence encoding, but rather an issue of representation entanglement, where evidence and priors are severely imbalanced during fusion.

Current mitigation methods primarily rely on contrastive decoding~\cite{vcd,avcd}, modality-adaptive generation~\cite{mad}, or preference optimization~\cite{omnidpo,moddpo}. As illustrated in Fig.~\ref{fig1}a, these methods treat hallucinations as output-level anomalies. They align cross-modal cues by calibrating final decoding distributions or optimizing response preferences. However, whether applied during training or inference, they inherently operate at the output level. In early-fusion OLLMs, grounding evidence and learned priors become deeply entangled long before decoding starts. Because these methods merely manipulate the final answer without controlling internal representations, they struggle to repair the corrupted hidden states that actually cause the hallucination, leaving the underlying conflict between grounding evidence and learned priors unresolved.

To address this issue, we propose \textbf{Omni-Woodpecker (OWP)}, a training-free framework that shifts output-level hallucination mitigation to representation-level rebalancing, without modifying model parameters or the language head. As illustrated in Fig.~\ref{fig:method_pipeline}, OWP follows a three-stage pipeline. It first identifies cross-modal conflicts and determines the target evidence together with its competing source from complementary modality-specific views. OWP then separates grounding evidence from competing priors to derive representation edits that preserve faithful evidence while removing misleading interference. Finally, it adaptively rebalances the fused representations according to the detected conflict, restoring suppressed evidence and attenuating deceptive priors before decoding.

Beyond introducing a new intervention pipeline, OWP relies on a fundamental principle: hallucinations should be corrected where they originate instead of where they appear. Rather than modifying the final decoding distribution, OWP explicitly identifies cross-modal conflicts and restores the balance between grounding evidence and learned priors directly in the hidden representation (\textit{i.e.}, restoring suppressed evidence while selectively attenuating misleading priors). As a result, the fused representation is rebalanced before decoding, leading to more reliable, evidence-driven reasoning. 

In summary, our main contributions are:
\begin{itemize}
    \item \textbf{Evidence-Prior Rebalancing Formulation.} We reformulate omni-modal hallucination as an internal imbalance between grounding evidence and learned priors within fused representations, revealing that hallucinations arise from representation entanglement rather than perception failure. This shifts the mitigation focus from output-level manipulation to representation-level repair.
    \item \textbf{Conflict-Conditioned Representation Rebalancing.} We propose OWP, a training-free framework that diagnoses query-conditioned cross-modal conflicts and selectively rebalances fused multimodal representations through evidence-preserving interventions, restoring suppressed evidence and attenuating misleading priors without modifying any model parameter.
    \item \textbf{Extensive Experimental Validation.} Experiments on CMM and AVHBench show that OWP outperforms the strongest baseline by \textbf{4.9} and \textbf{7.7} points and improves the original OLLMs by up to \textbf{12.9} points on hallucination-centric metrics. OWP further generalizes across different OLLM architectures and consistently improves performance on diverse omni-modal reasoning benchmarks.
\end{itemize}

Section~II reviews related work, Section~III presents OWP, Section~IV
reports experiments and analyses, Section~V covers limitations and future
work, and Section~VI concludes.

%% file: sec/02relatedwork.tex
\section{Related Work}
\subsection{Omni-Modal Models and Cross-Modal Hallucination}
OLLMs connect acoustic and visual encoders~\cite{clip,whisper} to a language
model. They build on audio-visual interfaces~\cite{videollama,salmonn} and
any-to-any architectures~\cite{zhan2024anygpt,nextgpt,imagebind}. They support
video understanding~\cite{han2024onellm,seth2026exploring,fu2026video}, spoken
interaction~\cite{ye2024cat,vita15}, and cross-modal
reasoning~\cite{videollama2,qwen25omni,qwen3omni,omnivinci,han2026omni}. Early
fusion enables joint reasoning, but it also lets an irrelevant sound, image, or
language prior override the evidence that a query needs. CMM describes these
failures as language, visual, and audio dominance and as spurious cross-modal
correlations. AVHBench tests audio-driven visual hallucination, visual-driven
audio hallucination, and audio-visual matching~\cite{cmm,avhbench}. The frozen
model often answers correctly from the query-relevant stream alone, even when
the fused input follows a competing cue (Table~\ref{tab:unified_ablation}).
This motivates edits that keep the encoded evidence rather than replace the
output.

\subsection{Output-Level Hallucination Mitigation}
Hallucination in vision--language models was first measured with
caption-based and object-probing benchmarks~\cite{chair,pope}. Later
diagnostic suites also cover mixed language--visual failures~\cite{mme,hallusionbench}.
One line of work treats hallucination as an error in the output distribution
and corrects it without touching the fused representation. Contrastive decoding
penalizes tokens that stay likely when the perceptual input is degraded. VCD
contrasts original and distorted visual inputs~\cite{vcd}. OPERA adds an
over-trust penalty and reallocates attention in hindsight~\cite{opera}. DoLa
improves factuality by contrasting logits from different layers~\cite{dola}.
Other variants contrast against perturbed instructions, visual grounding
states, or the model's own signals~\cite{icd,m3id,halc}. For audio-visual
models, AVCD estimates which of the three modalities dominates and contrasts
masked and full inputs~\cite{avcd}. MAD predicts whether a query needs audio,
video, or both, and weights its contrastive branches
accordingly~\cite{mad}. Post-hoc pipelines such as Woodpecker detect and revise
unsupported content after generation~\cite{woodpecker}. A second line changes
the model itself. Factually augmented RLHF and AI feedback improve grounding in
vision--language models~\cite{llavarlhf,rlaifv}. OmniDPO and MoD-DPO use
preference learning for perceptual alignment and for robustness to irrelevant
modalities~\cite{omnidpo,moddpo}. Both lines act on the decoding distribution
or on the model weights. In early-fusion OLLMs, however, evidence and priors
are already mixed before decoding starts. Neither line controls the
interaction that causes the error, and neither identifies which stream
competes with a given query.

\subsection{Representation-Level Intervention}
A more recent line edits internal activations directly. Model-editing methods
locate and change factual associations inside the model~\cite{rome,memit}, and
activation steering shifts behavior along chosen directions~\cite{actadd}.
Inference-time intervention finds truth-related directions and shifts
activations along them during generation~\cite{iti}. Representation
engineering derives such directions from contrasting prompt
sets~\cite{repe}. Contrastive activation addition steers a model with the
difference between two contrasting behaviors~\cite{caa}. \mbox{HulluEdit}
applies this idea to visual hallucination. It separates visual evidence from an
orthogonal anti-prior subspace and edits the residual state in one
pass~\cite{hulluedit}. These studies show that the factors behind
hallucination can be reached in the representation, not only in the output.
However, they assume a fixed split, usually perceptual evidence against a
language prior, and apply one direction to every input. Query-conditioned
methods such as MAD, in turn, use modality relevance only to weight output
branches. OWP instead finds, for each input, the stream that the query needs
and the stream that competes with it. It builds the edit directions from these
two streams and splits one correction budget between attention paths and the
residual state according to the diagnosed conflict.

%% file: sec/03method.tex
\section{Method}

\begin{figure*}[t]
\centering
\includegraphics[width=0.95\textwidth]{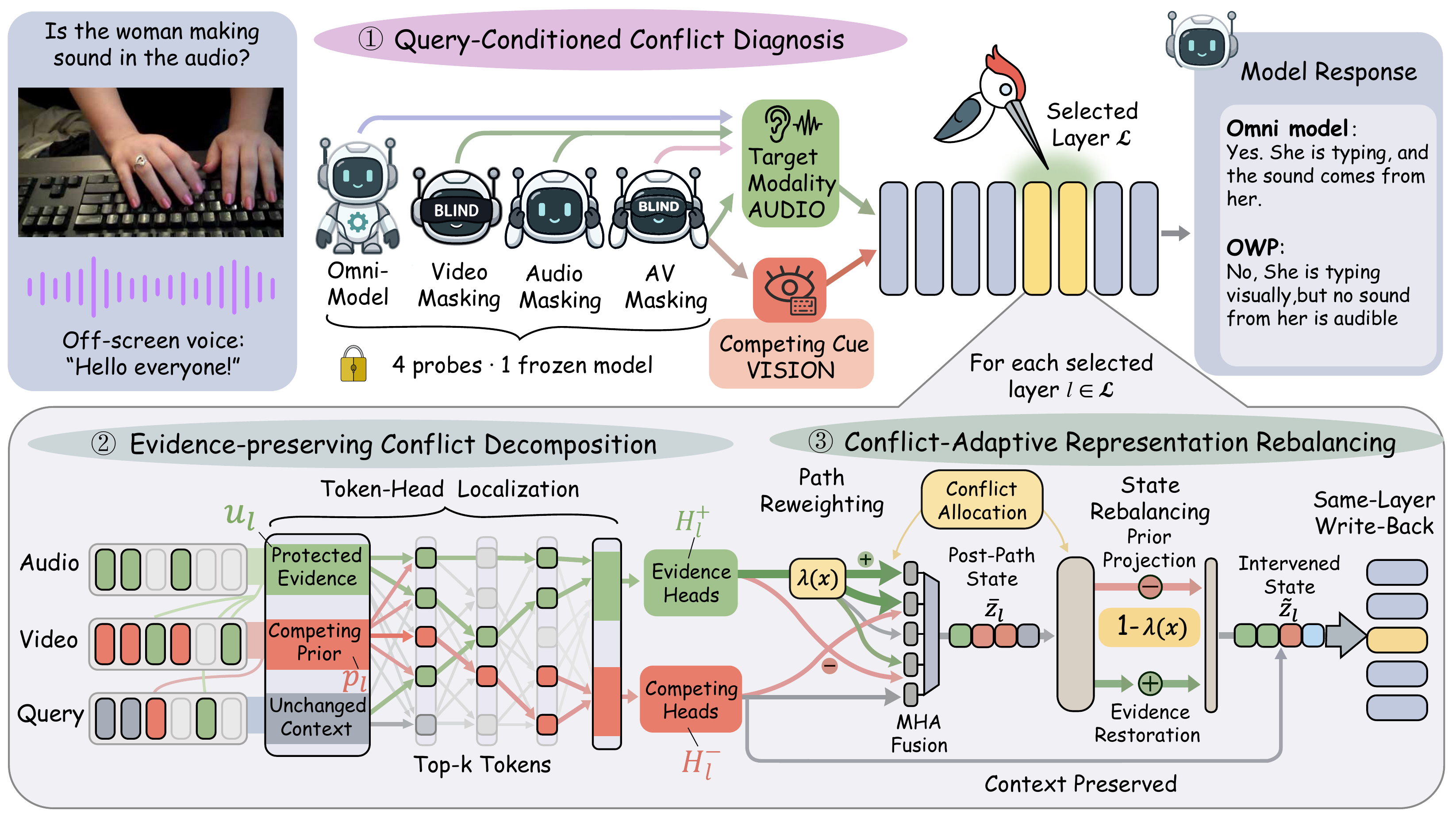}
\caption{Overview of Omni-Woodpecker (OWP). Given a query and an audio-visual input, OWP runs four views of a frozen model to find the target modality and the competing cue. It then splits token--head responses into protected evidence, competing prior, and unchanged context. Finally, it divides one correction budget between path reweighting and residual-state restoration and writes the edited state back to the same layer. Valid context is kept, and the cue that causes the hallucination is suppressed.}
\label{fig:method_pipeline}
\end{figure*}

\subsection{Overview}
OWP edits the fused representation of a frozen OLLM in three stages
(Fig.~\ref{fig:method_pipeline}). \textit{Query-conditioned conflict
diagnosis} compares several input views to find the stream that a query needs
and the stream that competes with it. \textit{Evidence-preserving conflict
decomposition} turns these view differences into a protected evidence
direction, an orthogonal competing direction, and a measured evidence deficit.
\textit{Conflict-adaptive representation rebalancing} then splits one
correction budget between attention-path edits and residual-state edits,
according to the diagnosed conflict.

\subsection{Query-Conditioned Conflict Diagnosis}
\label{sec:diagnosis}
The evidence that a query needs depends on the modality it asks about. OWP
therefore runs the same frozen model on four views: the full input
$x^{\mathrm F}=(q,v,a)$, the audio-only input $x^{\mathrm A}=(q,a)$, the
video-only input $x^{\mathrm V}=(q,v)$, and the text-only input
$x^{\mathrm T}=q$, where $q$, $v$, and $a$ are the query, video, and audio. A
\textit{cross-modal conflict} occurs when the modality that the query asks
about supports a different answer from the full view, while another modality
agrees with the full view. OWP first selects the target view from the
modality that the query asks about. It then picks the competing view from the
remaining inputs with two signals: agreement with the full-view answer and
source-normalized attention~\cite{mad,avcd}. The text-only view serves as a
shared reference that separates modality evidence from language-driven
answers. The pair $(x^{\mathrm{tar}},x^{\mathrm{non}})$ gives the evidence
source to protect and the competing source to suppress. No step uses labels
or reference answers.

\paragraph{Target typing}
We append a diagnostic query that asks which modality is needed: audio, video,
or both. We take the softmax $(\pi_{\mathrm A},\pi_{\mathrm V},\pi_{\mathrm{AV}})$
over the three next-token logits and form the scores
$\rho_{\mathrm A}=\pi_{\mathrm A}+\pi_{\mathrm{AV}}/2$ and
$\rho_{\mathrm V}=\pi_{\mathrm V}+\pi_{\mathrm{AV}}/2$. The higher score wins
if the gap exceeds a fixed margin; otherwise, a fixed fallback order applies
(Section~\ref{sec:impl}).

\paragraph{Competing-source verification}
The competing source is checked on the full view. For every layer except the
last, we sum the head-averaged attention of the final query token over the
audio, visual, and language spans. Each sum is divided by the span length and
averaged over layers, which gives one attention mass per source. A source is
accepted as competing only when two conditions hold. First, its single-stream
branch agrees with the full answer, while the target branch does not. Second,
its attention mass clearly exceeds that of the target. The ranking adds this
mass gap to the confidence score. When the cross-modal score and the
background-language score are too close to separate, OWP skips the edit.

\subsection{Evidence-Preserving Conflict Decomposition}

\paragraph{Calibrated evidence direction}
The non-target source may also carry useful evidence, so it is not treated as
harmful as a whole. OWP first finds the cross-view components that support the
target-view answer. Let $y^{\mathrm{tar}}$ and $y^{\mathrm F}$ be the
first-token predictions of the target view and the full view, and let $S_b$ be
the frozen logit score of view $b$. The preference of view $b$ between the two
answers is $\Delta_b=S_b(y^{\mathrm{tar}})-S_b(y^{\mathrm F})$. These
preferences give two non-negative weights, $w^{\mathrm{non}}$ and
$w^{\mathrm{fus}}$, shared by all edited layers. Let $\bar\Delta$ be the mean
preference of the non-target branch over these layers. Its size sets the pair
strength $\kappa=\tanh(|\bar\Delta|/\tau_s)$, and its sign sets the support
for each answer, $\pi_{\mathrm{tar}}=\sigma(\bar\Delta/\tau_s)$ and
$\pi_{\mathrm F}=\sigma(-\bar\Delta/\tau_s)$, where $\sigma$ is the sigmoid
and $\tau_s$ is a temperature. Thus, $\kappa$ shows how clearly the two answers
are separated, and $\pi_{\mathrm{tar}}$ and $\pi_{\mathrm F}$ show which answer
the non-target branch supports. The agreement between the target and full
branches gives two trust terms:
\begin{equation}
\begin{aligned}
t_{\mathrm{tar}}&=\left[\sigma(\Delta_{\mathrm{tar}}/\tau_s)\sigma(\Delta_{\mathrm F}/\tau_s)\right]^{1/2},\\
t_{\mathrm F}&=\left[\sigma(-\Delta_{\mathrm{tar}}/\tau_s)\sigma(-\Delta_{\mathrm F}/\tau_s)\right]^{1/2}.
\end{aligned}
\label{eq:trust_terms}
\end{equation}
These terms gate the weights as
$w^{\mathrm{non}}=\gamma\kappa\pi_{\mathrm{tar}}t_{\mathrm{tar}}$ and
$w^{\mathrm{fus}}=\gamma\kappa\pi_{\mathrm F}t_{\mathrm F}$, with a fixed gain
$\gamma$. A cross-view component is kept only when both branches support it.

Let $z_\ell^b$ be the last query-token state at layer $\ell$ for
$b\in\{\mathrm F,\mathrm{tar},\mathrm{non},\mathrm T\}$. We subtract the
text-only state to remove the query and the shared language component, and we
define $\mathcal N(\xi)=\xi/(\lVert\xi\rVert_2+\epsilon)$ with $\epsilon>0$.
The evidence direction is
\begin{equation}
u_\ell=\mathcal N\!\left[\mathcal N(e_\ell^{\mathrm{tar}})
    +w^{\mathrm{non}}\mathcal N(e_\ell^{\mathrm{non}})
    +w^{\mathrm{fus}}\mathcal N(e_\ell^{\mathrm{fus}})\right],
\label{eq:evidence_direction}
\end{equation}
where \(e_\ell^{\mathrm{tar}}=z_\ell^{\mathrm{tar}}-z_\ell^{\mathrm T}\),
\(e_\ell^{\mathrm{non}}=z_\ell^{\mathrm{non}}-z_\ell^{\mathrm T}\), and
\(e_\ell^{\mathrm{fus}}=z_\ell^{\mathrm F}-z_\ell^{\mathrm{tar}}\) are the
target evidence, the non-target information, and the fusion shift. The
direction $u_\ell$ is anchored on target evidence and keeps only the parts of
the other views that support the target. The inner normalization removes scale
differences between the terms, and the outer one gives a unit direction.

\paragraph{Orthogonal conflict direction and evidence deficit}
The fusion shift $e_\ell^{\mathrm{fus}}$ can hold both useful evidence and
competing content. With the evidence direction fixed, OWP separates the two and
measures how much target evidence the fused state has lost:
\begin{equation}
\begin{aligned}
p_\ell&=\mathcal N\!\left[e_\ell^{\mathrm{fus}}
    -(u_\ell^\top e_\ell^{\mathrm{fus}})u_\ell\right],\\
g_\ell&=\max(u_\ell^\top
    (z_\ell^{\mathrm{tar}}-z_\ell^{\mathrm F}),0).
\end{aligned}
\label{eq:conflict_decomposition}
\end{equation}
The direction $p_\ell$ is what remains after removing the part along the
evidence direction, so $p_\ell^\top u_\ell=0$. Suppressing $p_\ell$ thus leaves
the evidence component untouched. The clipped projection $g_\ell$ measures the
target evidence missing from the fused state, and it is zero when nothing is
missing. Together, \eqref{eq:evidence_direction} and
\eqref{eq:conflict_decomposition} decide what to protect, suppress, and restore.

\paragraph{Head-level conflict localization}
The directions tell OWP what to edit, and head localization tells it where.
OWP forms a competing head set $\mathcal H_\ell^-$ and an evidence head set
$\mathcal H_\ell^+$ from source attention and from the answer-aligned value
contributions $v_{\ell j}^{\mathrm{src}}$ and $v_{\ell j}^{\mathrm{evi}}$. At
each edited layer, OWP groups tokens by source and keeps the most-attended
tokens in each group. A head joins $\mathcal H_\ell^-$ only if it attends more
to the non-target group \emph{and} its value contribution favors the full-view
answer over the target-view answer. The set $\mathcal H_\ell^+$ uses the
opposite test and excludes $\mathcal H_\ell^-$. Both sets are ranked by
answer-aligned value contribution under a fixed cap per layer. An update is
skipped when either contribution is numerically unstable. The heads and values
are therefore chosen for each input, not fitted in advance.

\subsection{Conflict-Adaptive Representation Rebalancing}
\label{sec:method_rebalance}
The decomposition can be applied at two sites. Head-level edits change what
each source sends before aggregation, while residual edits change the
competition already merged into the hidden state. OWP uses the full-view answer
preference $\Delta_{\mathrm F}(x)$ to split the correction between the two
sites. For an identified conflict, the path share is
$\lambda(x)=\sigma((\tau-\Delta_{\mathrm F}(x))/T)$, where $\tau$ is the
allocation threshold and $T$ is the temperature. A weak preference gives a
larger $\lambda(x)$ and more control over what each source sends. A strong
preference moves the correction to the residual state through $1-\lambda(x)$.
The split thus matches the edit site to how far the conflict has progressed.

Let $o_\ell$ and $\widetilde o_\ell$ be the original and edited attention
outputs at layer $\ell$, and let $\bar z_\ell$ be the state after the path
edit. The coefficients $\alpha_{\mathrm P}$, $\beta_{\mathrm P}$,
$\alpha_{\mathrm R}$, $\beta_{\mathrm R}$, and $\eta_{\mathrm P}$ set path-level
evidence enhancement, path-level prior suppression, state-level evidence
restoration, state-level prior suppression, and path strength:
\begin{equation}
\begin{aligned}
\widetilde o_\ell
&=o_\ell+\eta_{\mathrm P}\lambda(x)\left[
-\beta_{\mathrm P}\sum_{j\in\mathcal H_\ell^-}v_{\ell j}^{\mathrm{src}}
+\alpha_{\mathrm P}\sum_{j\in\mathcal H_\ell^+}v_{\ell j}^{\mathrm{evi}}
\right],\\
\widetilde z_\ell
&=\bar z_\ell+[1-\lambda(x)]\left[
-\beta_{\mathrm R}(\bar z_\ell^\top p_\ell)p_\ell
+\alpha_{\mathrm R}g_\ell u_\ell\right].
\end{aligned}
\label{eq:hybrid_rebalancing}
\end{equation}
The head-level edit subtracts source-aligned values from $\mathcal H_\ell^-$
and adds evidence-aligned values to $\mathcal H_\ell^+$ before aggregation. The
residual edit removes the component along $p_\ell$ and restores the measured
deficit $g_\ell u_\ell$. The shares $\lambda(x)$ and $1-\lambda(x)$ divide one
correction budget between the two sites. The edit sites and coefficients are
fixed globally, while the directions, head sets, and allocation weights are
computed for each input.

\paragraph{Runtime order and decoding}
The subscripts $\mathrm R$ and $\mathrm P$ denote the residual branch and the
token--head path branch, not transformer layers. OWP runs only when a conflict
appears. It first runs the full view and the target view on the frozen model.
If their answers agree, there is no conflict, and OWP returns the backbone
answer. Otherwise, the non-target and text-only views give the competing
source, the edit directions, and the allocation weight, without choosing any
output token. For an identified conflict, one edited full-view pass follows.
At each selected layer, the path branch reweights token--head routes before
aggregation. The residual branch suppresses the orthogonal prior, restores the
measured deficit, and writes the result back to the same layer. This schedule
gives the same outputs as running all four views on every input. The language
head and the decoding rule are unchanged. Quantitative evaluation scores the
answer-option tokens with a finite-answer probe, and qualitative examples use
the frozen model's own autoregressive decoder.

\subsection{Implementation Details}
\label{sec:impl}
All constants are global and fixed before evaluation. Target typing uses a
margin of $0.12$ on $|\rho_{\mathrm A}-\rho_{\mathrm V}|$. Below this margin,
ties are broken by terms in the question, then by branch disagreement, and
finally by the backbone answer. The diagnostic query has four phrasings with
the same meaning and the same three response tokens
(Table~\ref{tab:typing_prompts}). Competing-source
verification uses a $0.05$ confidence margin and a $0.03$ margin between the
cross-modal and background scores. Evidence calibration uses $\tau_s{=}1.0$
and $\gamma{=}0.75$. Head localization keeps the 16 most-attended tokens in
each source group and at most four heads per layer. On \mbox{Qwen2.5-Omni-7B}, OWP edits evidence layers 12,
16, and 20 and prior layers 24, 26, and 27, weighted 0.25/0.35/0.40 and 0.23/0.61/0.16,
with
$\alpha_{\mathrm P}{=}\alpha_{\mathrm R}{=}0.10$,
$\beta_{\mathrm P}{=}\beta_{\mathrm R}{=}0.95$, $\eta_{\mathrm P}{=}0.25$,
$\tau{=}{-}9/16$, and $T{=}0.10$. Transfer to another backbone adapts
only the edited layers and two residual gains.

%% file: sec/04experiments.tex
\providecommand{\owpvisualizationfig}{Figures/qs.pdf}
\providecommand{\owpcheck}{}
\renewcommand{\owpcheck}{\ding{51}}
\providecommand{\owpcross}{}
\renewcommand{\owpcross}{\ding{55}}

\section{Experiments}
\label{sec:experiments}
We evaluate OWP on four questions: whether it corrects cross-modal conflicts,
whether it keeps joint audio-visual evidence, whether it transfers to another
backbone, and whether it preserves general omni-modal reasoning. We then ablate
its design and analyze where, and on which inputs, the edit changes the
backbone's predictions.

\begin{table*}[t]
\caption{Main results on CMM and AVHBench (\%). PA/HR: CMM perception
accuracy/hallucination resistance. F1: AVHBench official binary F1. All rows, including external backbones run
from released checkpoints, follow the protocol of Section~\ref{sec:setup}. Bold:
best within each group. Parentheses: overall change against the group backbone
(against \mbox{Qwen2.5-Omni-7B} for external models).}
\label{tab:main_results}
\centering
\small
\begin{tabular*}{\textwidth}{@{\extracolsep{\fill}}lcccc@{\hspace{12pt}}c@{\hspace{12pt}}ccccc@{\hspace{12pt}}c}
\toprule
\multirow{5}{*}{\textbf{Model / Method}} & \multicolumn{6}{c}{\textbf{CMM}} &
\multicolumn{5}{c}{\textbf{AVHBench}} \\
\cmidrule(lr){2-7}\cmidrule(lr){8-12}
& \textbf{Visual} & \textbf{Audio} & \textbf{Lang.}
& \multicolumn{3}{c}{\textbf{Overall}}
& \shortstack{\textbf{Audio}\\\textbf{$\to$Video}} & \shortstack{\textbf{Video}\\\textbf{$\to$Audio}} & \shortstack{\textbf{AV}\\\textbf{Match.}}
& \multicolumn{2}{c}{\textbf{Overall}} \\
\cmidrule(lr){5-7}\cmidrule(lr){11-12}
& \textbf{Acc.} & \textbf{Acc.} & \textbf{Acc.}
& \textbf{Acc.} & \textbf{PA} & \textbf{HR}
& \textbf{Acc.} & \textbf{Acc.} & \textbf{Acc.} & \textbf{Acc.} & \textbf{F1} \\
\midrule
\multicolumn{12}{@{}l}{\textit{External omni backbones}} \\
\mbox{VITA-1.5}
& 52.0 & 71.5 & 56.8 & 60.1 ($-13.1$) & 78.3 & 41.8
& 68.6 & 57.9 & 47.9 & 56.7 ($-18.4$) & 60.6 \\
OmniVinci
& 81.8 & \textbf{82.3} & 78.5 & 80.8 ($+7.6$) & 86.8 & \textbf{74.8}
& 70.7 & 69.7 & 64.0 & 67.9 ($-7.2$) & 68.6 \\
Qwen3-Omni
& \textbf{83.3} & 79.3 & \textbf{82.8} & \textbf{81.8} ($+8.6$) &
\textbf{91.3} & 72.2 & \textbf{84.0} & \textbf{80.1} &
\textbf{70.4} & \textbf{77.5} ($+2.4$) & \textbf{79.7} \\
\midrule
\multicolumn{12}{@{}l}{\textit{Qwen2.5-Omni-7B}} \\
Backbone
& 68.8 & 69.8 & 81.3 & 73.2 & 89.5 & 57.0
& 80.4 & 73.1 & 74.5 & 75.1 & 76.3 \\
AVCD
& 75.8 & 72.0 & 76.5 & 74.8 ($+1.6$) & 83.5 & 66.0
& 81.3 & 81.1 & 63.2 & 74.8 ($-0.3$) & 72.0 \\
MAD
& 76.8 & 84.5 & \textbf{82.5} & 81.2 ($+8.0$) & \textbf{89.7} & 72.8
& 84.6 & 78.3 & 68.0 & 76.0 ($+0.9$) & 75.0 \\
\mbox{HulluEdit}
& 75.5 & 81.5 & 80.8 & 79.2 ($+6.0$) & 86.2 & 72.3
& 83.8 & 82.2 & 51.4 & 71.6 ($-3.5$) & 66.4 \\
\textbf{OWP}
& \textbf{86.0} & \textbf{90.2} & 82.0 & \textbf{86.1 ($+12.9$)} & 86.7 &
\textbf{85.5} & \textbf{84.9} & \textbf{85.3} & \textbf{81.1} &
\textbf{83.7 ($+8.6$)} & \textbf{83.3} \\
\midrule
\multicolumn{12}{@{}l}{\textit{VideoLLaMA2-AV}} \\
Backbone
& 77.0 & 77.8 & 74.0 & 76.2 & 87.2 & 65.3
& 78.2 & 77.3 & 56.6 & 70.2 & 72.9 \\
AVCD
& 79.3 & 81.0 & 73.5 & 77.9 ($+1.7$) & 85.0 & 70.8
& \textbf{80.6} & 78.4 & 59.2 & 72.1 ($+1.9$) & \textbf{73.9} \\
MAD
& \textbf{83.3} & 77.0 & \textbf{77.3} & 79.2 ($+3.0$) & \textbf{88.0} & 70.3
& 79.6 & 76.3 & 56.6 & 70.0 ($-0.2$) & 71.7 \\
\mbox{HulluEdit}
& 79.3 & 79.3 & 73.0 & 77.2 ($+1.0$) & 82.2 & 72.2
& 78.3 & 78.6 & 59.3 & 71.7 ($+1.5$) & 70.1 \\
\textbf{OWP}
& 78.5 & \textbf{88.8} & 75.0 & \textbf{80.8 ($+4.6$)} & 83.8 &
\textbf{77.7} & 78.7 & \textbf{79.4} & \textbf{62.2} & \textbf{74.1 ($+3.9$)} &
\textbf{73.9} \\
\bottomrule
\end{tabular*}

\end{table*}

\subsection{Experimental Setup}
\label{sec:setup}
\paragraph{Benchmarks and baselines}
We evaluated all 1,200 CMM conflict cases and all 5,302 AVHBench examples,
including AV Matching~\cite{cmm,avhbench}, and the official splits of four
general omni-modal reasoning benchmarks~\cite{omnibench,worldsense,avqa,musicavqa}.
The controlled comparisons apply three baselines to the frozen
\mbox{Qwen2.5-Omni-7B} and \mbox{VideoLLaMA2-AV}
backbones~\cite{qwen25omni,qwen25,videollama2,avcd,mad,hulluedit}: AVCD for
contrastive decoding, MAD for modality-adaptive decoding, and \mbox{HulluEdit}
for representation editing. MAD uses a contrast coefficient of 2.5 on CMM and
0.5 on AVHBench, and AVCD uses 2.5. \mbox{HulluEdit} uses evidence rank 6,
prior rank 4, and a prior coefficient of 0.3. \mbox{VITA-1.5}, OmniVinci, and
Qwen3-Omni differ in architecture and training data~\cite{vita15,omnivinci,qwen3omni},
so we report them separately as references for architecture and scale. We ran
their released checkpoints on the same examples with the same prompts,
decoding, and scoring.

\paragraph{Protocol and statistics}
All controlled methods share the examples, media preprocessing, and frozen
backbones. Videos are sampled at 2 frames per second with at most 32 frames,
and the Qwen visual frontend uses between 100,352 and 602,112 pixels. OWP uses
no reference answers and no per-example supervision. Its edited layers come
from causal localization, and all coefficients and margins are fixed before
evaluation. Inputs that OWP skips keep the backbone output. CMM overall
accuracy is the mean of PA and HR, and AVHBench uses its official accuracy and
F1. Table~\ref{tab:main_results} uses the native protocol, with each benchmark's
own answer format, whereas Table~\ref{tab:unified_ablation} uses a shared binary
yes/no executor,
so backbone accuracies differ between the two. For \mbox{VideoLLaMA2-AV} in
Table~\ref{tab:main_results}, the evidence layers (16, 20, 24) and the residual
gains ($\alpha_{\mathrm R}{=}0.20$, $\beta_{\mathrm R}{=}1.20$) are adapted to
the backbone. Paired comparisons use the 4,604 examples on which every method
returns a valid prediction, with exact two-sided McNemar tests and 95\%
confidence intervals from 5,000 row-level bootstrap resamples. Mechanism
controls use 20,000 resamples with Holm correction within each family, and
random-direction controls use three fixed seeds.

\subsection{Main Results}
The main results test whether OWP corrects conflicts in both directions while
keeping the evidence that the query needs.

\paragraph{Conflict identification}
OWP edited only the inputs that contained a conflict. Under the native
protocol, on the CMM and directional AVHBench rows with complete view records, it identified a conflict
in $25.8\%$ of inputs and changed the final answer in $16.9\%$, and the other
$74.2\%$ ran the unmodified backbone. AV Matching is excluded here because it
has no single target modality. Target typing (Section~\ref{sec:diagnosis})
agreed with the benchmark-defined target modality on $99.48\%$ of rows. Only
8 of the 24 disagreeing rows reached an edit, with no net effect on
accuracy, so the gating is reliable.

\paragraph{Cross-benchmark performance}
OWP achieved the highest overall accuracy among the controlled methods on both
benchmarks (Table~\ref{tab:main_results}). With \mbox{Qwen2.5-Omni-7B}, it
reached 86.1\% on CMM and 83.7\% on the complete AVHBench, exceeding the
strongest baseline, MAD, by 4.9 and 7.7 points. On the common subset of CMM and
directional AVHBench, it outperformed the backbone, MAD, and \mbox{HulluEdit}
by 10.40, 4.69, and 3.52 points, and every paired difference was significant
($p<10^{-16}$). One intervention with fixed settings thus
corrects several conflict types on one backbone.

\paragraph{Bidirectional correction}
OWP improved both conflict directions and all three dominance types. It
reached 86.0\% and 90.2\% under CMM visual and audio dominance, and 84.9\% and
85.3\% on the two AVHBench directions. It also improved language dominance,
where the competing cue is the query text rather than another sensory stream
(81.3\% to 82.0\%). On CMM overall, OWP raised hallucination resistance (HR)
from 57.0\% to 85.5\% with perception accuracy (PA) at 86.7\%. This reduced the
backbone's 32.5-point gap between PA and HR to 1.2 points, whereas MAD kept a
16.9-point gap (89.7\% PA, 72.8\% HR). OWP thus brings perception and
hallucination resistance into near balance.

\paragraph{Joint audio-visual evidence}
OWP kept the evidence of both streams on AV Matching, which asks whether the
audio and the video belong together. Because this task has no single target
stream, OWP treats the joint audio-visual input as the evidence source and the
text-only view as the competing prior. It builds the evidence direction from
counterfactual inputs in which the audio or the video is replaced by another
clip. OWP reached 81.1\%, exceeding the backbone, MAD, and AVCD by 6.6, 13.1,
and 17.9 points. AVCD showed the opposite trade-off, with 81.2\% average
directional accuracy but only 63.2\% on AV Matching. OWP combined 84.9\% and
85.3\% directional accuracy with 81.1\% matching accuracy.

\paragraph{Cross-backbone transfer}
The decomposition transferred to a backbone with a different audio-visual
fusion design. On \mbox{VideoLLaMA2-AV}, OWP raised CMM accuracy from 76.2\% to
80.8\% and AVHBench accuracy from 70.2\% to 74.1\%, the best overall results in
that group. It also raised HR from 65.3\% to 77.7\% and AV Matching from 56.6\%
to 62.2\%. Moreover, OWP on \mbox{Qwen2.5-Omni-7B} exceeded the newer
Qwen3-Omni under the same protocol by 4.3 and 6.2 points on CMM and AVHBench,
without any training.
\begin{table}[t]
\caption{General omni-modal reasoning accuracy (\%) of \mbox{Qwen2.5-Omni-7B} with and without OWP. Parentheses: gain over the backbone.}
\label{tab:general_reasoning}
\centering
\footnotesize
\setlength{\tabcolsep}{2.5pt}
\newcommand{\gain}[1]{\scriptsize{(#1pp)}}
\begin{tabular}{@{}lc@{\hspace{10pt}}c@{\hspace{10pt}}c@{\hspace{10pt}}c@{}}
\toprule
\textbf{Model} & \textbf{OmniB.} & \textbf{WorldS.} & \textbf{MUSIC-AVQA} & \textbf{AVQA} \\
\midrule
Qwen2.5-Omni-7B & 45.36 & 45.21 & 85.08 & 91.83 \\
$+$ OWP & \shortstack{\textbf{47.29}\\\gain{+1.93}} &
\shortstack{\textbf{47.46}\\\gain{+2.25}} &
\shortstack{\textbf{86.08}\\\gain{+1.00}} &
\shortstack{\textbf{92.50}\\\gain{+0.67}} \\
\bottomrule
\end{tabular}

\end{table}

\begin{table}[t]
\caption{Inference cost per input: latency, latency relative to the backbone,
and forward calls. OWP uses the conflict-triggered schedule, timed per
component.}
\label{tab:inference_efficiency}
\centering
\small
\setlength{\tabcolsep}{3pt}
\begin{tabular}{@{}lrrrrr@{}}
\toprule
\textbf{Method} & \multicolumn{3}{c}{\shortstack{\textbf{Qwen2.5-}\\\textbf{Omni-7B}}} & \multicolumn{2}{c}{\textbf{VideoLLaMA2-AV}} \\
\cmidrule(lr){2-4}\cmidrule(lr){5-6}
& \textbf{Lat. (s)} & \textbf{Rel.} & \textbf{Calls}
& \textbf{Lat. (s)} & \textbf{Rel.} \\
\midrule
Backbone & 3.54 & 1.00$\times$ & 1 & 1.40 & 1.00$\times$ \\
MAD & 9.96 & 2.81$\times$ & 9 & 3.29 & 2.36$\times$ \\
\mbox{HulluEdit} & 13.02 & 3.68$\times$ & 5 & 1.94 & 1.38$\times$ \\
\textbf{OWP} & \textbf{6.05} & \textbf{1.71}$\times$ & $\leq$5 & \textbf{2.48} & \textbf{1.77}$\times$ \\
\bottomrule
\end{tabular}

\end{table}

\paragraph{Capability preservation}
OWP improved general omni-modal reasoning on all four benchmarks
(Table~\ref{tab:general_reasoning}). With the same frozen settings, accuracy
rose by 1.93 points on OmniBench, 2.25 on WorldSense, 1.00 on MUSIC-AVQA, and
0.67 on AVQA. All four benchmarks have closed answer sets, and OWP scores the
answer options with the same finite-answer probe. The largest gains appeared on
OmniBench and WorldSense, whose questions combine several streams. Correcting
cross-modal hallucination thus also strengthens general omni-modal reasoning.

\paragraph{Inference cost}
OWP was cheaper than both compared editing and decoding baselines on
\mbox{Qwen2.5-Omni-7B} (Table~\ref{tab:inference_efficiency}). Latency was
measured per input on an RTX A6000 and averaged over three runs. Under the
conflict-triggered schedule (Section~\ref{sec:method_rebalance}), the extra
views and the edited pass run only when the full and target views disagree. The
average latency was therefore $1.71\times$ and $1.77\times$ that of the backbone
on \mbox{Qwen2.5-Omni-7B} and \mbox{VideoLLaMA2-AV}, with at most five forward
calls per input. MAD required $2.81\times$ and $2.36\times$ with nine calls. On
\mbox{Qwen2.5-Omni-7B}, OWP was also more than twice as fast as
\mbox{HulluEdit} (6.05\,s against 13.02\,s). Peak memory was 24.9\,GiB and
17.3\,GiB on the two backbones.

\subsection{Ablation Studies}
\label{sec:ablation}
We ablate OWP under two protocols. Table~\ref{tab:mechanism_ablation} varies
the evidence and prior components, the edit sites, and the allocation strategy
under the native protocol of Table~\ref{tab:main_results}.
Table~\ref{tab:unified_ablation} applies one shared executor and one
configuration to both backbones, so component and view ablations can be
compared on the same rows under one answer protocol.
\begin{table}[t]
\caption{Ablation of OWP components, intervention sites, and allocation
strategies on \mbox{Qwen2.5-Omni-7B} (protocol of
Table~\ref{tab:main_results}). \owpcheck{}/\owpcross{}: enabled/disabled. AVH:
directional AVHBench accuracy. Bold: best per column.}
\label{tab:mechanism_ablation}
\centering
\scriptsize
\setlength{\tabcolsep}{1pt}
\begin{tabular}{@{}lccccc@{\hspace{5pt}}c@{\hspace{5pt}}c@{\hspace{5pt}}c@{}}
\toprule
\textbf{Variant} & \textbf{Evi.} & \textbf{Pri.} & \textbf{Res.} & \textbf{Pth.} & \textbf{Ada.} & \textbf{AVH} & \textbf{PA} & \textbf{HR} \\
\midrule
Backbone      & \owpcross & \owpcross & \owpcross & \owpcross & \owpcross & 75.51 & \textbf{89.50} & 57.00 \\
Evidence & \owpcheck & \owpcross & \owpcheck & \owpcheck & \owpcheck & 80.92 ($+5.41$) & 83.67 ($-5.83$) & 70.00 ($+13.00$) \\
Prior    & \owpcross & \owpcheck & \owpcheck & \owpcheck & \owpcheck & 82.23 ($+6.72$) & 85.33 ($-4.17$) & 84.33 ($+27.33$) \\
Residual & \owpcheck & \owpcheck & \owpcheck & \owpcross & \owpcross & 83.23 ($+7.72$) & 85.67 ($-3.83$) & 84.33 ($+27.33$) \\
Path     & \owpcheck & \owpcheck & \owpcross & \owpcheck & \owpcross & 81.68 ($+6.17$) & 84.67 ($-4.83$) & 69.67 ($+12.67$) \\
Fixed    & \owpcheck & \owpcheck & \owpcheck & \owpcheck & \owpcross & 83.93 ($+8.42$) & 84.67 ($-4.83$) & 82.33 ($+25.33$) \\
\midrule
\textbf{OWP} & \owpcheck & \owpcheck & \owpcheck & \owpcheck & \owpcheck & \textbf{85.17 ($+9.66$)} & 86.70 ($-2.80$) & \textbf{85.50 ($+28.50$)} \\
\bottomrule
\end{tabular}

\end{table}
\paragraph{Evidence enhancement and prior suppression}
Each component improved the backbone on its own, and the two together gave the
best result (Table~\ref{tab:mechanism_ablation}). Evidence enhancement raised
AVHBench accuracy by $5.41$ points and CMM HR by $13.00$ points. Prior
suppression raised them by $6.72$ and $27.33$ points. Combining the two
reached the best AVHBench accuracy ($85.17\%$) and HR ($85.50\%$) and cut the
PA cost from $5.83$ and $4.17$ points to $2.80$. The two components thus
offset each other's perception cost and outperform either alone.

\paragraph{Residual and path interventions}
The residual state was the main site where the competing cue was resolved.
With both components active, residual-only editing reached 83.23\% AVHBench
accuracy and 84.33\% HR, whereas path-only editing reached 81.68\% and
69.67\%, a 14.66-point gap in HR. Enabling both sites with a fixed split raised
AVHBench accuracy from 83.23\% to 83.93\% but lowered HR from 84.33\% to
82.33\%. A split that is identical for every input thus trades HR for accuracy,
which motivates a split that follows the conflict of each input.

\paragraph{Adaptive allocation and layer assignment}
Adaptive allocation improved on fixed allocation in all three metrics: AVHBench
accuracy rose from 83.93\% to 85.17\%, PA from 84.67\% to 86.70\%, and HR from
82.33\% to 85.50\% ($+1.24$, $+2.03$, and $+3.17$ points). The total budget is
unchanged, because $\lambda(x)$ and $1-\lambda(x)$ only redistribute it. When
the full view already strongly prefers the competing answer, $\lambda(x)$ moves
the budget to the residual branch, the site that the previous comparison found
most effective. Layer placement also mattered: swapping the layers of the two
components lowered accuracy by $6.79$ points (Holm-adjusted $p<0.001$).

\begin{table}[t]
\caption{Input-view ablation (shared executor, binary yes/no probe) on all
CMM and directional AVHBench rows with a single target modality. The frozen
backbone is re-run on the indicated stream with the unchanged prompt; the oracle
block uses the benchmark-defined target modality. The shift row adds a constant
to the yes-minus-no margin to match OWP's yes rate. OWP edits the full input.
Parentheses: paired difference against the full input (pp).}
\label{tab:unified_ablation}
\centering
\footnotesize
\setlength{\tabcolsep}{4pt}
\begin{tabular}{@{}lcc@{}}
\toprule
\textbf{Input} & \shortstack{\textbf{Qwen2.5-}\\\textbf{Omni-7B}}
& \textbf{VideoLLaMA2-AV} \\
\midrule
Full (backbone) & 82.06 & 76.85 \\
\midrule
\multicolumn{3}{@{}l}{\textit{One physical stream, same for every row}}\\
Audio only & 69.71 ($-12.35$) & 69.91 ($-6.94$) \\
Video only & 71.42 ($-10.64$) & 66.15 ($-10.70$) \\
Text only & 52.88 ($-29.18$) & 53.61 ($-23.24$) \\
\midrule
\multicolumn{3}{@{}l}{\textit{Stream chosen per row by the oracle target modality}}\\
Target stream only (oracle) & 84.28 ($+2.23$) & 77.56 ($+0.71$) \\
Competing stream only (oracle) & 56.85 ($-25.21$) & 54.50 ($-22.35$) \\
\midrule
Shift toward ``No'' (matched) & 82.66 ($+0.60$) & 77.52 ($+0.67$) \\
\midrule
\textbf{$+$ OWP} & \textbf{84.91 ($+2.85$)} & \textbf{78.99 ($+2.14$)} \\
\bottomrule
\end{tabular}

\end{table}

\begin{table}[t]
\caption{The four phrasings of the diagnostic query used for target typing.
Each phrasing is appended to the original query and keeps the same three
response tokens.}
\label{tab:typing_prompts}
\centering
\footnotesize
\setlength{\tabcolsep}{3pt}
\renewcommand{\arraystretch}{1.08}
\begin{tabular}{@{}c p{0.86\columnwidth}@{}}
\toprule
\textbf{ID} & \textbf{Diagnostic instruction appended to the query} \\
\midrule
1 & \textit{Identify which modality is required to answer the question (audio, video, or both).} \\
2 & \textit{Given this question, select the necessary modality for reasoning (audio, video, or both).} \\
3 & \textit{Which modality does this question require (audio, video, or both)?} \\
4 & \textit{State the modality relevant for answering this question (audio, video, or both).} \\
\bottomrule
\end{tabular}
\end{table}

\paragraph{One shared configuration across backbones}
A single configuration, fixed on \mbox{Qwen2.5-Omni-7B}, improved both
backbones without any backbone-specific change. This shared executor sets
$\alpha_{\mathrm R}{=}0.10$ and $\beta_{\mathrm R}{=}0.95$, edits layers 12,
16, 20, 24, 26, and 27, and keeps the same head-localization limits and
orthogonal protection. It scores each view with one binary yes/no logit over
all CMM and directional AVHBench rows. Under this executor, OWP edited 11.33\% of these rows on
\mbox{Qwen2.5-Omni-7B} and 20.90\% on \mbox{VideoLLaMA2-AV}, and every other row
matched the backbone output exactly. The shared configuration improved the two
backbones by $+2.85$ and $+2.14$ points (Table~\ref{tab:unified_ablation}),
with media-clustered 95\% confidence intervals of $[1.95, 3.75]$ and
$[1.41, 2.88]$. The gain remained positive on both backbones when either
residual coefficient was changed by $\pm20\%$. The gain was concentrated on rows with a
negative ground truth ($+5.53$ and $+5.97$ points), while positive rows changed
by at most $1.7$ points ($+0.17$ and $-1.69$). The edit therefore removes
unsupported ``Yes'' answers, the most common form of hallucination in these
yes/no benchmarks.

\paragraph{Answer-prior control}
The gain of OWP is not a uniform bias toward ``No''. We shifted the backbone's
yes-minus-no margin toward ``No'' by a constant that matched OWP's yes rate
(52.0\% and 55.4\%). This shift gained only $+0.60$ and $+0.67$ points
(Table~\ref{tab:unified_ablation}), far below OWP (exact McNemar $p<10^{-6}$
and $p<10^{-3}$). On \mbox{Qwen2.5-Omni-7B}, the shift also lowered
positive-row accuracy from 86.73\% to 85.39\%, whereas OWP raised it to
86.90\%.

\paragraph{Component necessity}
Under the shared executor, prior suppression supplied most of the correction,
and evidence restoration amplified it. Keeping only evidence enhancement (prior
coefficient set to zero) changed accuracy by $+0.04$ and $-0.02$ points.
Keeping only prior suppression (evidence coefficient set to zero) changed it by
$+1.34$ and $+1.51$ points, whereas the full model gained $+2.85$ and $+2.14$
points. With the allocation fixed, the full model therefore gained more than
the two single components combined on both backbones. In
Table~\ref{tab:mechanism_ablation}, by contrast, the adaptive schedule gives
each single component the whole budget.

\paragraph{View necessity}
The competing stream, not the loss of information, degraded the backbone's
answer (Table~\ref{tab:unified_ablation}). No single stream matched the full
input: audio alone cost $12.35$ and $6.94$ points, video alone $10.64$ and
$10.70$, and text alone $29.18$ and $23.24$, within four points of the $50.0\%$
chance level of these balanced probes. Choosing the stream for each row with
the oracle target modality separates the two effects. Keeping only the
competing stream sharply lowered accuracy ($-25.21$ and $-22.35$), whereas
keeping only the target stream raised it ($+2.23$ and $+0.71$). OWP infers the
target online and keeps every modality in the input, yet it exceeded this
oracle routing on both backbones ($84.91\%$ against $84.28\%$ and $78.99\%$
against $77.56\%$).

\subsection{Mechanism Analysis}
\label{sec:mechanism}
\begin{figure*}[t]
\centering
\includegraphics[width=1.0\textwidth]{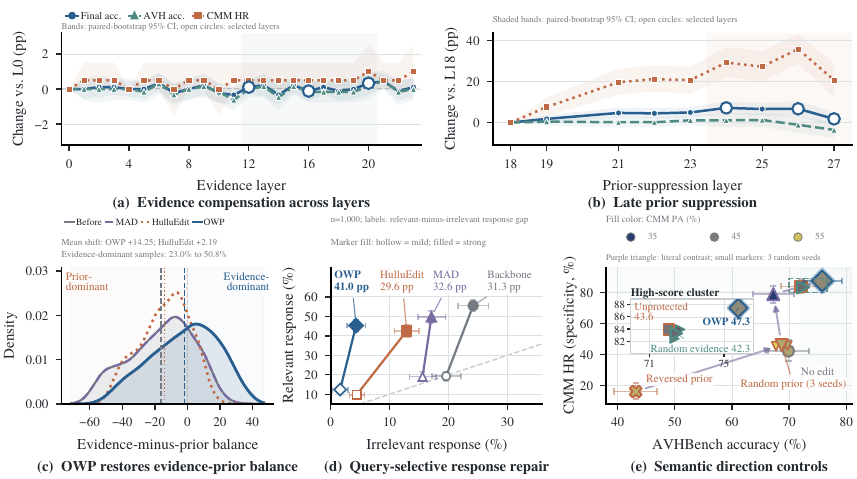}
\caption{Mechanism analysis of OWP. (a,b) Change in overall accuracy, AVH
(directional AVHBench) accuracy, and CMM HR when evidence enhancement (a) or
prior suppression (b) is applied at a single layer; bands show paired-bootstrap
95\% CIs, and open circles mark the layers that OWP edits. (c) Evidence-minus-prior
balance before (frozen backbone) and after each method; positive values denote
evidence-dominant states. (d) Answer-change rate when the query-relevant
(vertical) or irrelevant (horizontal) stream is corrupted (1,000 samples); hollow
and filled markers denote mild and strong corruption, and points farther above
the dashed $y{=}x$ line respond more selectively. (e) AVHBench accuracy against
CMM HR for OWP and direction controls on the conflict subset that OWP edits;
fill encodes CMM PA, error bars show row-bootstrap 95\% CIs, and small markers
show three projection-matched random seeds. Only OWP, with both semantic
directions intact, reaches the high-accuracy, high-HR region.}
\label{fig:completed_mechanism}
\end{figure*}

We next analyze where the two components act, how they change the hidden
representation, whether the edits follow meaningful directions, and how OWP
responds when relevant or irrelevant streams are corrupted. Unless stated
otherwise, these analyses use the conflict subset of CMM and AVHBench, that is,
the inputs that OWP actually edits. On these inputs, the target-view answer
differs from the backbone answer, and the PA and HR in
Fig.~\ref{fig:completed_mechanism}e are computed on this subset. The corruption
analysis uses its own matched set of inputs.

\paragraph{Layer-wise intervention analysis}
The two components acted at different depths
(Fig.~\ref{fig:completed_mechanism}a,b). Evidence enhancement was effective
across a broad range of middle layers, so query-relevant evidence can be
recovered from several neighboring layers. Prior suppression became effective
only in later layers, and in the full single-layer scan its effect rose
steadily toward a late peak. This order matches the two operations: evidence is
best recovered while it is still separable, whereas the competing prior can be
removed only after it has formed. Together with the layer-swap control of
Section~\ref{sec:ablation}, these profiles support evidence recovery in middle
layers and prior suppression in later layers.

\paragraph{Changes in hidden representations}
OWP moved the fused representation toward the evidence. It shifted the
evidence-minus-prior balance by 14.25 points, compared with 2.19 points for
\mbox{HulluEdit} (Fig.~\ref{fig:completed_mechanism}c), and raised the share of
evidence-dominant states from 23.0\% to 50.8\%. Across the three
prior-suppression layers, the prior coordinate fell steadily, while the
evidence coordinate changed by at most 0.14 points. OWP therefore reduces the
prior and leaves the evidence direction intact.

\paragraph{Behavioral selectivity under modality corruption}
OWP made the answer follow the queried stream. We corrupted either the queried
stream or the other stream on 1,000 matched inputs. Mild corruption mutes or
blacks out alternating one-second segments, and strong corruption removes the
stream. Selectivity is the gap between answer changes under relevant and
irrelevant corruption. The gap was $41.0$ points for OWP, against $31.3$ for the
backbone, $32.6$ for MAD, and $29.6$ for \mbox{HulluEdit}
(Fig.~\ref{fig:completed_mechanism}d). Under strong corruption, OWP changed its
answer for 45.3\% of relevant corruptions but for only 4.3\% of irrelevant
ones. This selectivity follows directly from deriving each edit from the
queried view (Section~\ref{sec:diagnosis}).

\paragraph{Semantic direction controls}
The edits worked through their directions rather than their size
(Fig.~\ref{fig:completed_mechanism}e). Reversing the prior direction sharply
lowered both axes, and random prior directions with matched projection stayed
well below OWP across three seeds. The evidence direction set the operating
point: replacing it with a random direction lowered AVHBench accuracy and CMM
HR relative to OWP ($72.5\%$ and $83.9\%$ against $75.8\%$ and $87.4\%$), and
removing orthogonal protection had a similar effect ($72.0\%$ and $83.9\%$).
OWP reached high accuracy and high HR only with both directions intact. The
prior direction removes the competing cue from the fused representation, and
the evidence direction keeps the edited state on relevant content.

\subsection{Qualitative Results}
\begin{figure*}[t]
\centering
\includegraphics[width=1.0\textwidth]{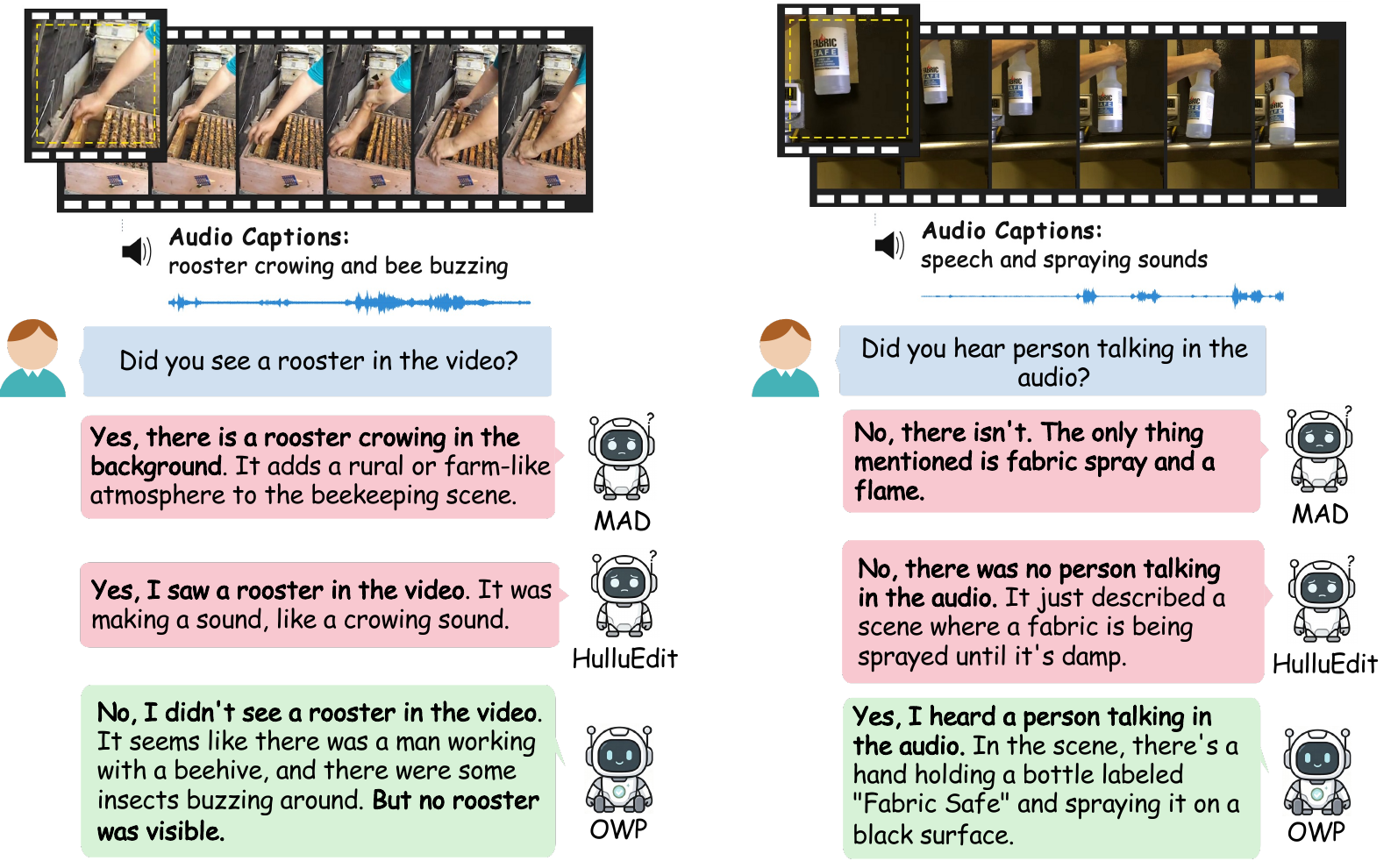}
\caption{Qualitative comparison on reciprocal visual- and audio-grounded queries;
pink and green boxes mark incorrect and correct answers. Left: given audible
crowing, MAD and \mbox{HulluEdit} report a visually absent rooster. Right: given a
visual spraying scene, they deny the concurrent speech. OWP answers both from the
queried modality while keeping compatible context.}
\label{fig:qualitative_visualization}
\end{figure*}

Fig.~\ref{fig:qualitative_visualization} shows that OWP answers from the
queried modality in both directions. In a beekeeping clip with audible crowing,
MAD and \mbox{HulluEdit} followed the audio cue and reported a rooster that is
not visible. In a clip showing fabric being sprayed, they followed the visual
stream and denied the speech heard at the same time. Each baseline followed the
more salient modality rather than the queried one. OWP answered both queries
correctly: it described the beehive without inventing the rooster, and
reported both the spraying and the speech.

%% file: sec/06limitations.tex
\section{Limitations and Future Work}

\subsection{Limitations}
OWP has four limitations that follow from its scope and design. First,
omni-modal hallucination covers several types of failure, and this paper
focuses on cross-modal hallucination, where the evidence from one modality is
overridden by a cue from another modality or from the language prior. CMM and
AVHBench are built around this failure, and OWP's diagnosis is designed for it.
Second, OWP restores evidence that the backbone encodes but fails to use during
fusion. It reads this evidence from the single-stream views of the frozen
model, so it does not add perception that the backbone lacks: when the target
modality is misperceived even on its own, there is no reliable evidence to
restore. Third, OWP edits hidden states and attention outputs, so it requires
access to model internals. Like other representation-level
methods~\cite{iti,hulluedit}, it applies to open-weight OLLMs but not to models
that are available only through an API. Fourth, like other inference-time
methods that contrast several input views~\cite{avcd,mad}, OWP requires extra
forward passes. Its conflict-triggered schedule keeps this cost at
$1.71$--$1.77\times$ the backbone latency, lower than that of MAD on both
backbones.

\subsection{Future Work}
The principle of OWP, which diagnoses the stream that a query needs and
rebalances evidence and priors inside the fused representation, extends
naturally to other classes of omni-modal hallucination. For temporal
hallucination, where a model misreports the order, duration, or synchrony of
audio-visual events, the diagnosis would locate when each stream provides
evidence and compare event timing across streams. For acoustic and
paralinguistic hallucination, where a model misattributes speaker identity,
emotion, prosody, or non-speech sounds, the diagnosis would separate spoken
content from acoustic attributes within the audio stream. Because OWP leaves
the weights unchanged, it can also be combined directly with training-based
methods. More broadly, our results show that cross-modal hallucination comes
from how evidence and priors interact inside the fused representation, and
correcting this internal evidence flow is a direct path toward reliable
omni-modal reasoning.

%% file: sec/05conclusion.tex
\section{Conclusion}
This paper presented Omni-Woodpecker (OWP), a training-free framework that
corrects cross-modal hallucination in OLLMs by rebalancing evidence and priors
inside the fused representation. OWP diagnoses the target and competing
streams from four views of a frozen model, separates an evidence direction from
an orthogonal prior direction, and splits one correction budget between
attention paths and the residual state. With \mbox{Qwen2.5-Omni-7B}, OWP
improved accuracy by 12.9 points on CMM and 8.6 points on AVHBench, raised CMM
hallucination resistance from 57.0\% to 85.5\% with perception accuracy at
86.7\%, and reached 81.1\% on AV Matching. It also exceeded the newer Qwen3-Omni by
4.3 and 6.2 points on the two benchmarks, and its edits made the answer follow
the queried modality, with a 41.0-point selectivity gap under stream
corruption. It transferred to
\mbox{VideoLLaMA2-AV} and improved all four general omni-modal reasoning
benchmarks by up to 2.25 points. Ablations and mechanism analyses showed that
prior suppression and evidence restoration are complementary, that the residual
state is the main site of correction, and that evidence is best restored in
middle layers and priors are best suppressed in later layers. Correcting
hallucination where it forms, inside the fused representation, is thus a
practical path toward reliable omni-modal reasoning without retraining.